\documentclass[conference]{IEEEtran}
\IEEEoverridecommandlockouts

\usepackage{cite}
\usepackage{amsmath,amssymb}
\usepackage{graphicx}
\usepackage{booktabs}
\usepackage{tikz}
\usetikzlibrary{positioning,arrows.meta,fit,calc}
\usepackage{multirow}
\usepackage{xcolor}
\usepackage{url}
\usepackage[hidelinks]{hyperref}
\expandafter\def\csname hbnum@add.gain.knn\endcsname{$+0.0104$}
\expandafter\def\csname hbnum@add.gain.profile\endcsname{$+0.0103$}
\expandafter\def\csname hbnum@audit.cnt.ge\endcsname{25{,}057}
\expandafter\def\csname hbnum@audit.coldtest.w\endcsname{Twelve}
\expandafter\def\csname hbnum@audit.kg.1hop.ends\endcsname{767{,}927}
\expandafter\def\csname hbnum@audit.kg.1hop.lineage\endcsname{278{,}358}
\expandafter\def\csname hbnum@audit.kg.1hop.wrong\endcsname{117{,}260}
\expandafter\def\csname hbnum@audit.kg.2hop.wrong\endcsname{99{,}312}
\expandafter\def\csname hbnum@audit.kg.publish.ends\endcsname{25{,}078}
\expandafter\def\csname hbnum@audit.kg.publish.ok\endcsname{4}
\expandafter\def\csname hbnum@audit.minint\endcsname{6}
\expandafter\def\csname hbnum@audit.num.items.diffmin\endcsname{25{,}077}
\expandafter\def\csname hbnum@audit.num.kgsets.w\endcsname{four}
\expandafter\def\csname hbnum@audit.num.sets.w\endcsname{six}
\expandafter\def\csname hbnum@audit.pop.cnt\endcsname{0.0320}
\expandafter\def\csname hbnum@audit.pop.train\endcsname{0.0358}
\expandafter\def\csname hbnum@audit.rel.1hop\endcsname{27}
\expandafter\def\csname hbnum@audit.rel.homo\endcsname{7}
\expandafter\def\csname hbnum@audit.rel.publish\endcsname{6}
\expandafter\def\csname hbnum@audit.share.max\endcsname{0.67}
\expandafter\def\csname hbnum@audit.share.med\endcsname{0.60}
\expandafter\def\csname hbnum@audit.share.min\endcsname{0.50}
\expandafter\def\csname hbnum@audit.sp.cnt\endcsname{0.80}
\expandafter\def\csname hbnum@audit.sp.train\endcsname{0.71}
\expandafter\def\csname hbnum@bd.model.d\endcsname{$+0.012$}
\expandafter\def\csname hbnum@bd.p1.ease\endcsname{0.006}
\expandafter\def\csname hbnum@bd.p1.tfg\endcsname{0.034}
\expandafter\def\csname hbnum@bd.p100.ci\endcsname{[$-0.004$, $-0.002$]}
\expandafter\def\csname hbnum@bd.p100.d\endcsname{$-0.003$}
\expandafter\def\csname hbnum@bd.p100.items\endcsname{1{,}214}
\expandafter\def\csname hbnum@bd.space.d\endcsname{$+0.005$}
\expandafter\def\csname hbnum@best.rep.r10\endcsname{0.1578}
\expandafter\def\csname hbnum@bud.best.max\endcsname{354}
\expandafter\def\csname hbnum@bud.best.min\endcsname{105}
\expandafter\def\csname hbnum@bud.cens.all.n.w.cap\endcsname{Five}
\expandafter\def\csname hbnum@bud.cens.allimpr.n.w.cap\endcsname{Four}
\expandafter\def\csname hbnum@bud.cens.max\endcsname{8}
\expandafter\def\csname hbnum@bud.cens.max.w\endcsname{eight}
\expandafter\def\csname hbnum@bud.cens.min\endcsname{7}
\expandafter\def\csname hbnum@bud.cens.min.w\endcsname{seven}
\expandafter\def\csname hbnum@bud.cens.s42.w\endcsname{eight}
\expandafter\def\csname hbnum@bud.cens.s43.w\endcsname{eight}
\expandafter\def\csname hbnum@bud.cens.s44.w\endcsname{seven}
\expandafter\def\csname hbnum@bud.cens.total\endcsname{29}
\expandafter\def\csname hbnum@bud.dsl44.best\endcsname{96}
\expandafter\def\csname hbnum@bud.dsl44.stop\endcsname{111}
\expandafter\def\csname hbnum@bud.gap1hop.100\endcsname{0.0613}
\expandafter\def\csname hbnum@bud.gap1hop.500\endcsname{0.0374}
\expandafter\def\csname hbnum@bud.kgrec1hop.d\endcsname{$+0.0262$}
\expandafter\def\csname hbnum@bud.kgrec1hop.pct\endcsname{31\%}
\expandafter\def\csname hbnum@bud.kgrecpubfix.100\endcsname{0.1204}
\expandafter\def\csname hbnum@bud.kgrecpubfix.500\endcsname{0.1379}
\expandafter\def\csname hbnum@bud.lgcl43.d\endcsname{$-0.0001$}
\expandafter\def\csname hbnum@bud.lgcn.100\endcsname{0.1297}
\expandafter\def\csname hbnum@bud.lgcn.500\endcsname{0.1362}
\expandafter\def\csname hbnum@bud.lgcn.d\endcsname{$+0.0065$}
\expandafter\def\csname hbnum@bud.lgcn.rng\endcsname{0.0048--0.0076}
\expandafter\def\csname hbnum@bud.maxpct\endcsname{31\%}
\expandafter\def\csname hbnum@bud.trunc.impr\endcsname{21}
\expandafter\def\csname hbnum@bud.trunc.pairs\endcsname{23}
\expandafter\def\csname hbnum@cc.a.flip.w\endcsname{ten}
\expandafter\def\csname hbnum@cc.a.pairs\endcsname{27}
\expandafter\def\csname hbnum@cc.a.stable.w\endcsname{eight}
\expandafter\def\csname hbnum@cc.b.flip.w\endcsname{one}
\expandafter\def\csname hbnum@cc.b.pairs.w\endcsname{ten}
\expandafter\def\csname hbnum@cc.b.stable.w\endcsname{eight}
\expandafter\def\csname hbnum@cc.c.flip.w\endcsname{one}
\expandafter\def\csname hbnum@cc.c.pairs\endcsname{36}
\expandafter\def\csname hbnum@cc.c.stable\endcsname{28}
\expandafter\def\csname hbnum@cls.all.f1lit\endcsname{0.404}
\expandafter\def\csname hbnum@cls.all.f1maj\endcsname{0.384}
\expandafter\def\csname hbnum@cls.all.field\endcsname{99.99\%}
\expandafter\def\csname hbnum@cls.best.rep\endcsname{0.885}
\expandafter\def\csname hbnum@cls.test.n\endcsname{27{,}345}
\expandafter\def\csname hbnum@cls.test.reported\endcsname{28{,}682}
\expandafter\def\csname hbnum@cold.new\endcsname{708}
\expandafter\def\csname hbnum@cold.parents.r10\endcsname{0.186}
\expandafter\def\csname hbnum@cold.removed\endcsname{12{,}705}
\expandafter\def\csname hbnum@cold.share.pct\endcsname{20\%}
\expandafter\def\csname hbnum@cold.test\endcsname{354}
\expandafter\def\csname hbnum@cold.tfidf.r10\endcsname{0.323}
\expandafter\def\csname hbnum@cold.tflin.r10\endcsname{0.354}
\expandafter\def\csname hbnum@cold.tflin.vs.tf\endcsname{$+0.031$}
\expandafter\def\csname hbnum@cold.tflin.vs.tf.ci\endcsname{[$+0.025$, $+0.037$]}
\expandafter\def\csname hbnum@cold.typed.vs.untyped\endcsname{$-0.0026$}
\expandafter\def\csname hbnum@cold.typed.vs.untyped.ci\endcsname{[$-0.0075$, $+0.0019$]}
\expandafter\def\csname hbnum@cold.users\endcsname{5{,}664}
\expandafter\def\csname hbnum@cold.withparent\endcsname{3{,}540}
\expandafter\def\csname hbnum@dan.r10\endcsname{0.1746}
\expandafter\def\csname hbnum@dan.vs.bestrep.pct\endcsname{11\%}
\expandafter\def\csname hbnum@dan.vs.ease\endcsname{$+0.0062$}
\expandafter\def\csname hbnum@dan.vs.ease.ci\endcsname{[$+0.0049$, $+0.0074$]}
\expandafter\def\csname hbnum@data.items\endcsname{25{,}080}
\expandafter\def\csname hbnum@data.items.dataset\endcsname{3{,}634}
\expandafter\def\csname hbnum@data.items.model\endcsname{16{,}200}
\expandafter\def\csname hbnum@data.items.space\endcsname{5{,}246}
\expandafter\def\csname hbnum@data.likes\endcsname{1{,}065{,}220}
\expandafter\def\csname hbnum@data.users\endcsname{38{,}624}
\expandafter\def\csname hbnum@ease.lambda\endcsname{200}
\expandafter\def\csname hbnum@ease.r10\endcsname{0.1684}
\expandafter\def\csname hbnum@ease.vs.lgcn\endcsname{$+0.0383$}
\expandafter\def\csname hbnum@ease.vs.lgcn.ci\endcsname{[$+0.0355$, $+0.0410$]}
\expandafter\def\csname hbnum@ex.smol.ent\endcsname{366541}
\expandafter\def\csname hbnum@ex.smol.item\endcsname{17166}
\expandafter\def\csname hbnum@ex.smol.ntrip\endcsname{153}
\expandafter\def\csname hbnum@ex.smol.nwrong.w\endcsname{one}
\expandafter\def\csname hbnum@ex.smol.wrongname\endcsname{\texttt{dpbos/postvalidation}}
\expandafter\def\csname hbnum@ex.vlm.ent\endcsname{268711}
\expandafter\def\csname hbnum@ext.time.s\endcsname{11}
\expandafter\def\csname hbnum@fd.L2.inter.o.hi\endcsname{$+0.0132$}
\expandafter\def\csname hbnum@fd.L2.inter.o.lo\endcsname{$+0.0103$}
\expandafter\def\csname hbnum@fd.L2.inter.s42\endcsname{$-0.0199$}
\expandafter\def\csname hbnum@fd.L2.rel.ep.hi\endcsname{33}
\expandafter\def\csname hbnum@fd.L2.rel.ep.lo\endcsname{27}
\expandafter\def\csname hbnum@fd.L3.red.max\endcsname{0.0279}
\expandafter\def\csname hbnum@fd.L3.red.min\endcsname{0.0188}
\expandafter\def\csname hbnum@fld.desc.n\endcsname{19{,}191}
\expandafter\def\csname hbnum@fld.desc.vs.ease\endcsname{$+0.0134$}
\expandafter\def\csname hbnum@fld.id.r10\endcsname{0.1850}
\expandafter\def\csname hbnum@fld.id.vs.full\endcsname{$+0.0005$}
\expandafter\def\csname hbnum@fld.id.vs.full.ci\endcsname{[$-0.0006$, $+0.0016$]}
\expandafter\def\csname hbnum@fld.mask.n\endcsname{18{,}251}
\expandafter\def\csname hbnum@fld.mask.vs.ease\endcsname{$+0.0125$}
\expandafter\def\csname hbnum@fld.meta.vs.ease\endcsname{$+0.0083$}
\expandafter\def\csname hbnum@fldg.id.r10\endcsname{0.1850}
\expandafter\def\csname hbnum@fldg.id.vs.full\endcsname{$-0.0010$}
\expandafter\def\csname hbnum@fldg.id.vs.full.ci\endcsname{[$-0.0021$, $+0.0001$]}
\expandafter\def\csname hbnum@grid.text\endcsname{\{3,5,10,30,100\}}
\expandafter\def\csname hbnum@kg.misaligned.w\endcsname{three}
\expandafter\def\csname hbnum@kg.models.m\endcsname{1.2 million}
\expandafter\def\csname hbnum@kg.released.w\endcsname{four}
\expandafter\def\csname hbnum@kgalign.publish\endcsname{4}
\expandafter\def\csname hbnum@kgbd.longtail.share\endcsname{54\%}
\expandafter\def\csname hbnum@kgbd.p1.fix\endcsname{0.020}
\expandafter\def\csname hbnum@kgbd.p1.rel\endcsname{0.004}
\expandafter\def\csname hbnum@kgbd.p100.fix\endcsname{0.155}
\expandafter\def\csname hbnum@kgbd.p100.rel\endcsname{0.148}
\expandafter\def\csname hbnum@kgbd.p20.fix\endcsname{0.040}
\expandafter\def\csname hbnum@kgbd.p20.rel\endcsname{0.011}
\expandafter\def\csname hbnum@kgbd2.p1.fix\endcsname{0.020}
\expandafter\def\csname hbnum@kgbd2.p1.rel\endcsname{0.004}
\expandafter\def\csname hbnum@kgbd2.p100.fix\endcsname{0.157}
\expandafter\def\csname hbnum@kgbd2.p100.rel\endcsname{0.133}
\expandafter\def\csname hbnum@kgcl.homo.gap\endcsname{0.0019}
\expandafter\def\csname hbnum@kgcl.publishfix.r10\endcsname{0.1604}
\expandafter\def\csname hbnum@kgin.o3.1hopfix.4\endcsname{0.1445}
\expandafter\def\csname hbnum@kgin.o3.publishfix\endcsname{0.1475}
\expandafter\def\csname hbnum@kgin.rel.1hop.r10\endcsname{0.0004}
\expandafter\def\csname hbnum@leak.clean.ci\endcsname{[$-0.00003$, $+0.00009$]}
\expandafter\def\csname hbnum@leak.clean.d\endcsname{$+0.00002$}
\expandafter\def\csname hbnum@leak.cnt.ci\endcsname{[$-0.00041$, $-0.00003$]}
\expandafter\def\csname hbnum@leak.cnt.d\endcsname{$-0.00021$}
\expandafter\def\csname hbnum@lgcl.s42.rq1\endcsname{0.1561}
\expandafter\def\csname hbnum@lgcl.s42.seeds\endcsname{0.1558}
\expandafter\def\csname hbnum@lr.lgcl.5.ep\endcsname{498}
\expandafter\def\csname hbnum@lr.lgcl.margin.min\endcsname{0.0021}
\expandafter\def\csname hbnum@mis.maxpct.all\endcsname{73\%}
\expandafter\def\csname hbnum@mis.minpct.all\endcsname{4\%}
\expandafter\def\csname hbnum@mt.diff.max\endcsname{0.0077}
\expandafter\def\csname hbnum@mt.diff.min\endcsname{0.0046}
\expandafter\def\csname hbnum@orig.kg.collected\endcsname{December 2024}
\expandafter\def\csname hbnum@orig.kg.edge.types\endcsname{30}
\expandafter\def\csname hbnum@orig.kg.edges\endcsname{6.2 million}
\expandafter\def\csname hbnum@orig.kg.nodes\endcsname{2.6 million}
\expandafter\def\csname hbnum@pc.cos.hi\endcsname{0.22}
\expandafter\def\csname hbnum@pc.cos.lo\endcsname{0.13}
\expandafter\def\csname hbnum@perm.gap.100\endcsname{0.0153}
\expandafter\def\csname hbnum@perm.gap.500\endcsname{0.0103}
\expandafter\def\csname hbnum@perm.indist.items\endcsname{8}
\expandafter\def\csname hbnum@perm.indist.users\endcsname{8}
\expandafter\def\csname hbnum@perm.runs.w\endcsname{nine}
\expandafter\def\csname hbnum@pex.ends.pct\endcsname{51.0}
\expandafter\def\csname hbnum@pex.gap.100\endcsname{0.0180}
\expandafter\def\csname hbnum@pex.gap.500\endcsname{0.0264}
\expandafter\def\csname hbnum@pex.items.pct\endcsname{98.4}
\expandafter\def\csname hbnum@pub.cos\endcsname{0.40}
\expandafter\def\csname hbnum@pub.dup\endcsname{15\%}
\expandafter\def\csname hbnum@r7.def.500\endcsname{0.1604}
\expandafter\def\csname hbnum@r7.r10.500\endcsname{0.1619}
\expandafter\def\csname hbnum@rep.hccf.absdelta.3\endcsname{0.002}
\expandafter\def\csname hbnum@rep.hccf.rep.hi\endcsname{0.1275}
\expandafter\def\csname hbnum@rep.hccf.rep.lo\endcsname{0.1211}
\expandafter\def\csname hbnum@rep.kgcl.2hop.d\endcsname{$-0.0015$}
\expandafter\def\csname hbnum@rep.kgrec.2hop.r10\endcsname{0.0942}
\expandafter\def\csname hbnum@rep.kgrec.2hop.rep\endcsname{0.0941}
\expandafter\def\csname hbnum@rep.lightgcn.nmatch.w\endcsname{eight}
\expandafter\def\csname hbnum@rep.within.all\endcsname{0.0010}
\expandafter\def\csname hbnum@rnd.cos\endcsname{0.03}
\expandafter\def\csname hbnum@rnd.r10\endcsname{0.0004}
\expandafter\def\csname hbnum@rpt.lightgcl\endcsname{0.0008}
\expandafter\def\csname hbnum@rq1.total.w\endcsname{eleven}
\expandafter\def\csname hbnum@seed.all.maxstd\endcsname{0.0073}
\expandafter\def\csname hbnum@seed.lgcl.s44.ep\endcsname{15}
\expandafter\def\csname hbnum@seed.lgcl.std\endcsname{0.0230}
\expandafter\def\csname hbnum@tf.r10\endcsname{0.1846}
\expandafter\def\csname hbnum@tf.vs.bestrep.pct\endcsname{17\%}
\expandafter\def\csname hbnum@tf.vs.ease\endcsname{$+0.0161$}
\expandafter\def\csname hbnum@tf.vs.ease.ci\endcsname{[$+0.0146$, $+0.0177$]}
\expandafter\def\csname hbnum@tfg.r10\endcsname{0.1860}
\expandafter\def\csname hbnum@tfg.vs.bestrep.pct\endcsname{18\%}
\expandafter\def\csname hbnum@tfg.vs.tf\endcsname{$+0.0014$}
\expandafter\def\csname hbnum@tfg.vs.tf.ci\endcsname{[$+0.0006$, $+0.0024$]}
\expandafter\def\csname hbnum@time.best.s\endcsname{358}
\expandafter\def\csname hbnum@time.ease.s\endcsname{253}
\expandafter\def\csname hbnum@time.kgin.onehop.h\endcsname{3.9}
\expandafter\def\csname hbnum@time.lightgcn.min\endcsname{25}
\expandafter\def\csname hbnum@tr.card.h1\endcsname{0.542}
\expandafter\def\csname hbnum@tr.card.mrr\endcsname{0.542}
\expandafter\def\csname hbnum@tr.hyb.mrr\endcsname{0.831}
\expandafter\def\csname hbnum@tr.hyb.vs.transe.hi\endcsname{$+0.2881$}
\expandafter\def\csname hbnum@tr.hyb.vs.transe.lo\endcsname{$+0.2746$}
\expandafter\def\csname hbnum@tr.mention.pct\endcsname{54.3\%}
\expandafter\def\csname hbnum@tr.transe.h1\endcsname{0.442}
\expandafter\def\csname hbnum@tr.transe.mrr\endcsname{0.549}
\expandafter\def\csname hbnum@tr.transe.mrr.hi\endcsname{0.558}
\expandafter\def\csname hbnum@tr.transe.mrr.lo\endcsname{0.544}
\expandafter\def\csname hbnum@tr.transe.rep\endcsname{0.559}
\expandafter\def\csname hbnum@trace.finetune.pct\endcsname{94\%}
\expandafter\def\csname hbnum@trace.n\endcsname{2{,}055}
\expandafter\def\csname hbnum@val.checks\endcsname{34}
\expandafter\def\csname hbnum@val.r.checks\endcsname{219}
\expandafter\def\csname hbnum@val.r.files\endcsname{40}
\expandafter\def\csname hbnum@val.r.mem\endcsname{1.8}
\expandafter\def\csname hbnum@val.r.time\endcsname{87}
\expandafter\def\csname hbnum@val.r.viol\endcsname{75{,}238}
\expandafter\def\csname hbnum@xr.mcclk.ent\endcsname{7{,}008}
\expandafter\def\csname hbnum@xr.mcclk.ent.doc\endcsname{182{,}011}

\newcommand{\tabnote}[1]{\par\vspace{3pt}\parbox{\linewidth}{\footnotesize #1}}
\newcommand{\num}[1]{\ifcsname hbnum@#1\endcsname\csname hbnum@#1\endcsname\else\textcolor{red}{\textbf{[??#1]}}\fi}

\begin{document}

\title{Auditing a Model-Hub Benchmark: Data Quality, Training Budgets,\\ and Side Information in HuggingBench}

\author{\IEEEauthorblockN{Micha{\l} Podstawski}
\IEEEauthorblockA{NASK National Research Institute\\
Warsaw, Poland\\
michal.podstawski@nask.pl}}

\maketitle

\begin{abstract}
Benchmarks built from model hubs combine interaction data, knowledge graphs (KGs) and resource documentation, but valid
files do not ensure that these artifacts describe the same objects. Treating the release as a data product, we audit
HuggingBench (SIGIR 2025), a benchmark for recommending Hugging Face resources, through data validation, a replication of
its \num{rq1.total.w} recommendation methods and controlled experiments with three training seeds. In \num{kg.misaligned.w}
of its \num{kg.released.w} external KGs, the files pass key and range checks but violate the item--entity identity
constraint on which KG recommenders rely. Correcting this join improves every KG-based method in every seed, by up to
\num{mis.maxpct.all} in mean Recall@10, and consistently reverses \num{cc.a.flip.w} of \num{cc.a.pairs} comparisons between
KG configurations. The protocol matters as well: the 100-epoch budget ends before early stopping in \num{bud.cens.min.w} to
\num{bud.cens.max.w} of \num{bud.cens.total} validation-selected settings, and \num{bud.trunc.impr} of \num{bud.trunc.pairs}
such setting--seed pairs improve with a longer budget. The released item text, which no reported method uses, is the
strongest signal: closed-form linear models with text and the item graph exceed the best reported Recall@10 by
\num{tfg.vs.bestrep.pct} while fitting in minutes on a CPU. Checks of other recommendation releases show that individually
valid releases of one dataset can use incompatible identifiers. We release corrected files, mappings, validators and logs,
and derive practices for preserving identity and provenance across the artifacts of benchmarks and model lakes.
\end{abstract}

\begin{IEEEkeywords}
data quality, benchmarking, model hubs, model lakes, provenance, recommender systems, reproducibility
\end{IEEEkeywords}

\section{Introduction}
Public model hubs have become the main distribution channel of machine-learning artifacts. A crawl of Hugging Face from
\num{orig.kg.collected} contains \num{kg.models.m} models~\cite{huggingbench}, together with datasets, papers and
interactive applications (Spaces), and a large share of the models are derivatives of popular base models: fine-tunes,
quantized variants, adapters and merges~\cite{hfreuse}. The data management community has begun to treat such collections as \emph{model lakes}, which call
for the same functionality as data lakes---search, provenance, versioning and quality control---applied to models and
their metadata~\cite{modellakes,modelhub,modeldb,mlflow}. One user-facing instance of this functionality is
recommendation: helping users find the models, datasets and Spaces that are relevant to them.

HuggingBench~\cite{huggingbench} turned this problem into a benchmark. It crawled HuggingKG, a knowledge graph of the
Hugging Face ecosystem with \num{orig.kg.nodes} nodes and \num{orig.kg.edges} edges, and released three test collections on
top of it: resource recommendation from user likes, task classification, and lineage tracing (predicting the base model of
a derived model). For recommendation, it released interaction splits and four external KGs that describe the items, and
reported \num{rq1.total.w} recommenders implemented in the SSLRec framework~\cite{sslrec}, from graph collaborative
filtering to KG-based recommendation.

A benchmark is a data product: its files, splits and protocol fix what every follow-up method will be compared on, and
errors in them propagate silently into the literature~\cite{labelerrors,recleakage,kapoor2023}. As the first benchmark for
its task, HuggingBench sets the reference point for future work. We therefore audit it end to end, asking
three questions that are central to data engineering for machine learning: \emph{Are the released data what they claim to
be?} \emph{Do the reported conclusions depend on undocumented properties of the data and the protocol?} \emph{Which signals
that the data already contain are left unused?} Our answers rest on checks of the released recommendation files and
implementations, a replication of all reported recommendation methods, and new experiments that vary one property at a
time; for the classification and tracing tasks, we examine the signals available in the released text.

\paragraph*{Findings} The reported results of the recommenders replicate closely (Section~\ref{sec:rep}), but the audit
changes what the benchmark's numbers mean in four ways.
\begin{itemize}
\item \emph{Misaligned identifiers} (Sections~\ref{sec:audit} and~\ref{sec:fix}). In \num{kg.misaligned.w} of the
  \num{kg.released.w} external KGs, items and the entities that describe them have different identifiers, so KG
  recommenders, which assume that item $i$ is entity $i$, attach the KG facts to the wrong items. The KG files pass key and
  range checks; a join by content reveals the error. Correcting the files
  improves every KG-based method on every graph in every seed, most strongly for rarely liked items, and reverses
  \num{cc.a.flip.w} of \num{cc.a.pairs} comparisons between KG configurations.
\item \emph{Truncated training} (Section~\ref{sec:budget}). Depending on the seed, the configuration selected on validation
  data exhausts the 100-epoch budget before early stopping in \num{bud.cens.min.w} to \num{bud.cens.max.w} of
  \num{bud.cens.total} settings; \num{bud.trunc.impr} of these \num{bud.trunc.pairs} truncated runs improve with a longer
  budget, which also shrinks the apparent cost of the misalignment.
\item \emph{Unused side information} (Section~\ref{sec:side}). The benchmark ships the text of every item (model and dataset
  cards, tags, metadata), but no reported method uses it. Closed-form linear models with this text achieve the highest
  scores we observed, and lexical TF-IDF representations beat dense embeddings from two modern encoders under matched
  conditions. Type and identifier alone score close to the full text, and descriptions remain useful under an identifier mask.
\item \emph{Provenance tasks} (Section~\ref{sec:prov}). Model cards name their base models: a training-free reading of the
  derived model's card nearly matches the best KG embedding method for lineage tracing. For newly published derivatives,
  text and lineage are complementary signals of user interest in a simulated cold-start setting.
\end{itemize}

\paragraph*{Contributions} Our contribution is an empirical study of integrity failures across independently exported
artifacts of a released benchmark: how to detect and repair them, and how they and the evaluation protocol change downstream
conclusions. (C1) An analysis of these failures in HuggingBench---identity mismatches between KG files that pass key and range
checks, six unmapped encodings of the same interactions, and features that may include held-out interactions---with
corrected KG files, the recovered mappings and executable checks, which pass on other releases in the same format but show that releases of one dataset can use
incompatible numberings. (C2) A replication of all
\num{rq1.total.w} recommendation methods and controlled experiments that isolate the effects of the data errors, the training
budget and the unused side information, over three seeds and with paired confidence intervals. (C3) Strong, cheap baselines and an inheritance
cold-start protocol for recommending new derivatives on model hubs. (C4) Lessons for builders of benchmarks and model lakes (Section~\ref{sec:lessons}).

\section{Background and Related Work}
\label{sec:related}
\paragraph*{Model hubs and model lakes} Systems for managing machine-learning models track versions, lineage and metadata of
models and experiments~\cite{modelhub,modeldb,mlflow}; the model lake vision extends data-lake functionality---discovery,
provenance, citation and quality control---to large public model collections~\cite{modellakes}. Empirical studies of Hugging
Face analyze how models are reused and maintained~\cite{hfreuse,hfevolution}, and recent work recovers lineage from model
weights~\cite{mother}. Lineage also helps predict the benchmark performance of new derivatives~\cite{tamura2025crow}; model
cards and metadata have been mined to reconstruct the supply chain of models and datasets~\cite{hugginggraph}; and graphs of
models, datasets and evaluations have been used to predict missing results, with a masking analysis of shortcuts in the
metadata~\cite{artifactlinker}. HuggingBench~\cite{huggingbench} complements these lines with a KG of the whole ecosystem and with
benchmarks for recommendation, classification and lineage tracing; it is, to our knowledge, the only benchmark for
recommending hub resources to users.

\paragraph*{Data quality for ML benchmarks} Data validation is an established part of ML pipelines: declarative constraints
and unit tests for data, including user-defined checks~\cite{deequ}, schema-based validation of training and serving
data~\cite{datavalidation}, and debugging of data distributions in ML pipelines~\cite{mlinspect}. We use the same principle of
executable checks and study one contract between independently exported artifacts---that an identifier denotes the same
object in every file that uses it---together with the consequences of violating it for a benchmark's conclusions.
Documentation standards ask dataset creators to state how data were collected and processed~\cite{datasheets,croissant}.
Errors in released benchmarks, such as label errors in test sets~\cite{labelerrors}, and leakage between training and
evaluation data~\cite{kapoor2023,recleakage} can change the ranking of methods. Knowledge-graph quality is usually assessed
for completeness and correctness of facts~\cite{kgrefinement}; the failure studied here concerns the identity link
between interaction data and otherwise well-formed KG files.

\paragraph*{Reproducibility in recommendation} Systematic studies found that many neural recommenders do not outperform
well-tuned simple baselines~\cite{dacrema2019,dacrema2021,anelli2022}, that conclusions depend on evaluation
details~\cite{sun2020} and on how much each method is tuned~\cite{rendle2020,shehzad2023}. Item-item linear
models~\cite{slim,ease} and their normalized variant DAN~\cite{dan} are strong top-$N$ recommenders, and closed-form
extensions add item side information to them~\cite{jeunen2020,l3ae}. Recent audits found inconsistencies between the
released artifacts of recommender papers and their descriptions~\cite{ferraridacrema2025artifact}, and showed that
third-party implementations of a recommender can diverge from the original~\cite{hidasi2023thirdparty}. We apply this
perspective to a new kind of item catalog, whose items come with technical documentation and explicit lineage, and to a
failure in the join between independently exported data files.

\section{HuggingBench and Experimental Setup}
\label{sec:setup}
\paragraph*{Data and protocol} HuggingKG was crawled from the Hugging Face API in \num{orig.kg.collected} and contains
models, datasets, Spaces, papers, collections, users and organizations connected by \num{orig.kg.edge.types} relation types.
The recommendation data are the 5-core subgraph of user likes of models, datasets and Spaces: \num{data.users} users and
\num{data.items} items (\num{data.items.model} models, \num{data.items.dataset} datasets, \num{data.items.space} Spaces) with
\num{data.likes} likes, split per user into 60\%/20\%/20\% for training, validation and testing. Methods rank all items,
excluding the user's training items, and are compared by Recall@$K$ and NDCG@$K$, $K\in\{5,10,20,40\}$. Configurations
are trained with early stopping on validation data: SSLRec evaluates every 3 epochs and stops after 5 evaluations without
improvement of its first metric, within a budget of 100 epochs. This criterion, named MRR@5 in SSLRec, sums the reciprocal
ranks of all relevant items in the top 5 before averaging over users; we keep it for selection and report it in
Fig.~\ref{fig:curves}. The external KGs are the 1-hop and
2-hop neighborhoods of the items in HuggingKG, a homogeneous item--item graph with seven relation types (fine-tune,
quantization, adapter, merge, model--training-dataset, Space--model, Space--dataset), and a publisher graph.

\paragraph*{Reported methods} The original study reports graph collaborative filtering (LightGCN~\cite{lightgcn},
HCCF~\cite{hccf}, SimGCL~\cite{simgcl}, LightGCL~\cite{lightgcl}, AutoCF~\cite{autocf}, DCCF~\cite{dccf}), social
recommendation (DSL~\cite{dsl}, MHCN~\cite{mhcn}), and KG-based recommendation (KGIN~\cite{kgin}, KGCL~\cite{kgcl},
KGRec~\cite{kgrec}) with each external KG. The best reported result is KGCL with the homogeneous graph (Recall@10
\num{best.rep.r10}); KGIN obtains near-zero scores in every setting.

\paragraph*{Implementations} We use SSLRec (commit \texttt{7597c31}) with the configuration files, data handlers and run
scripts of the HuggingBench repository, including its hyperparameter grids, on NVIDIA RTX~3090 GPUs; KGRec with three
layers on the 2-hop graphs needs more than 24~GB and runs on an RTX~5090 with otherwise identical code (with PyTorch 2.7,
which the RTX~5090 requires; a calibration epoch with two layers agreed with the RTX~3090 to within
$10^{-3}$ in validation MRR@5 and Recall@10). Running KGCL
required one compatibility fix (a private SciPy method removed in recent releases, replaced by an equivalent construction).
For KGIN, the framework's port does not train its embeddings on a GPU, an instance of the divergence of third-party
ports~\cite{hidasi2023thirdparty}; this defect is unrelated to the data, so we
evaluate KGIN with its authors' original implementation (commit \texttt{0e43517}), driven by the same data, evaluation and
early-stopping protocol; we use this implementation for all KGIN results in the analysis of the KGs. We select every
configuration of the reported methods on validation data, by the early-stopping criterion of the original study. All GPU runs use PyTorch 2.6 (CUDA 11.8) unless stated otherwise. Our additional methods are implemented in
NumPy/SciPy and run on an 8-core CPU; they use the same data, ranking over all items and masking of training items, but
their hyperparameters are selected differently, on a fixed quarter of the validation users by Recall@10: EASE $\lambda\in\{200,500,1000\}$; DAN user exponent $\beta\in\{0,0.25,0.5\}$,
$\lambda\in\{50,200,800\}$ and item exponent $\alpha\in\{0,0.1,\dots,0.4\}$; ItemKNN $k\in\{50,200\}$ and shrinkage
$\in\{0,10,100\}$; RP3beta $\alpha\in\{0.5,1\}$, $\beta\in\{0.2,0.4,0.6\}$, $k=200$; side-information and fusion
weights on small logarithmic grids (Section~\ref{sec:side}).

\paragraph*{Statistics} Every trained configuration of the analysis is run with three seeds (42, the seed of the released
configuration files, 43 and 44) and selected on validation data in each seed. Tables report the mean over seeds, except for
the replication rows, which report seed 42, and we call a
difference \emph{consistent} when it has the same sign in all three seeds; the per-seed values of every setting are in our
artifacts. Among validation-selected settings, the standard deviation of Recall@10 over seeds is at most
\num{seed.all.maxstd}, except for LightGCL (\num{seed.lgcl.std}), whose validation value in one seed stops improving after
epoch \num{seed.lgcl.s44.ep}; fixed configurations can vary more (Section~\ref{sec:budget}). Comparisons with a single fitted model per method, such as our deterministic baselines
against the reproduced LightGCN (seed 42), report 95\% confidence intervals of paired differences from a bootstrap over test
users with 1{,}000 resamples (for lineage tracing, over test triples); these intervals are conditional on the fitted models
and the fixed split and do not include training variance.

\section{Auditing the Released Data}
\label{sec:audit}
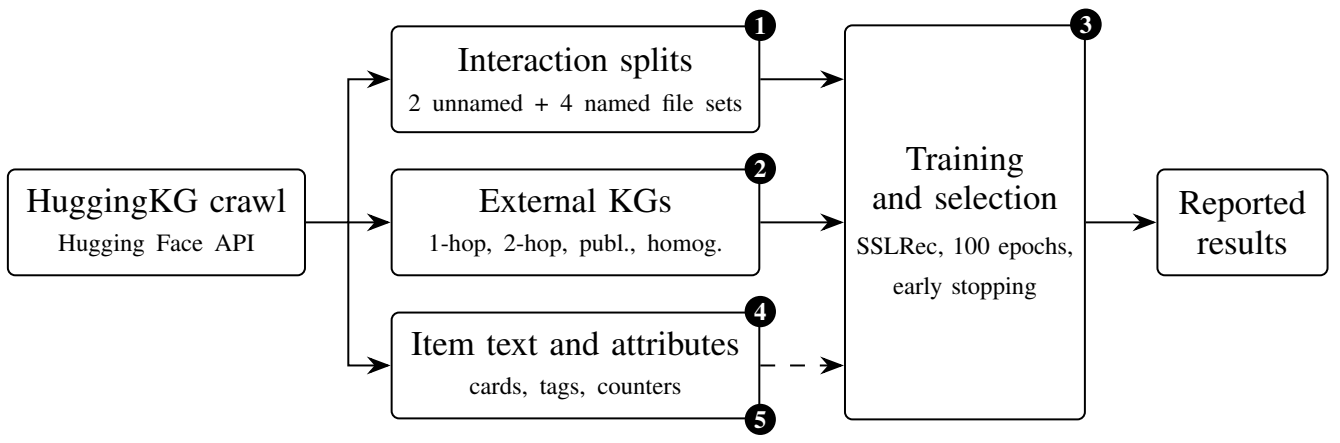
\begin{figure}[t]
  \centering
  \resizebox{\columnwidth}{!}{\begin{tikzpicture}[
  font=\scriptsize, >=Stealth,
  box/.style={draw, rounded corners=1.5pt, align=center, inner sep=1.5pt, fill=white},
  mid/.style={box, text width=25mm, minimum height=7.5mm},
  issue/.style={circle, draw, fill=black, text=white, inner sep=0.6pt, font=\tiny\bfseries},
  sub/.style={font=\tiny}]
  \node[mid] (ext) at (0,0) {External KGs\\{\tiny 1-hop, 2-hop, publ., homog.}};
  \node[mid, above=2.5mm of ext] (inter) {Interaction splits\\{\tiny 2 unnamed + 4 named file sets}};
  \node[mid, below=2.5mm of ext] (attr) {Item text and attributes\\{\tiny cards, tags, counters}};
  \node[box, text width=20mm, minimum height=7.5mm, left=6mm of ext] (kg) {HuggingKG crawl\\{\tiny Hugging Face API}};
  \path (inter.north east) ++(6mm,0) coordinate (t1);
  \path (attr.south east) ++(23mm,0) coordinate (t2);
  \node[box, fit=(t1)(t2), inner sep=0pt] (train) {};
  \node[align=center] at (train.center) {Training\\and selection\\[1pt]{\tiny SSLRec, 100 epochs,}\\{\tiny early stopping}};
  \node[box, text width=11mm, minimum height=7.5mm, right=5mm of train] (res) {Reported\\results};
  \coordinate (fork) at ($(kg.east)!0.5!(ext.west)$);
  \draw[->] (kg.east) -- (fork) |- (inter.west);
  \draw[->] (fork) -- (ext.west);
  \draw[->] (fork) |- (attr.west);
  \draw[->] (inter.east) -- (inter.east -| train.west);
  \draw[->] (ext.east) -- (ext.east -| train.west);
  \draw[->, dashed] (attr.east) -- (attr.east -| train.west);
  \draw[->] (train.east |- res.west) -- (res.west);
  \node[issue] at (inter.north east) {1};
  \node[issue] at (ext.north east) {2};
  \node[issue] at (attr.north east) {4};
  \node[issue] at (attr.south east) {5};
  \node[issue] at (train.north east) {3};
\end{tikzpicture}}
  \caption{The HuggingBench pipeline and the issues found by the audit: \textcircled{\scriptsize 1}~six encodings of
  the same interactions; \textcircled{\scriptsize 2}~item--entity misalignment in three external KGs;
  \textcircled{\scriptsize 3}~a training budget that ends before early stopping; \textcircled{\scriptsize 4}~crawl-time
  counters that may include held-out interactions; \textcircled{\scriptsize 5}~item text and attributes that contain the targets of
  the provenance tasks. Item text is released but used by no reported recommender (dashed).}
  \label{fig:pipeline}
\end{figure}

We checked the released recommendation files for internal consistency and against HuggingKG, and the reported
recommendation results against the protocol. The checks fall into six groups, each implemented as an executable test in our artifacts:
\begin{enumerate}
\item \emph{Split integrity}: splits are disjoint, the $k$-core property holds, and per-user proportions match the stated
  split.
\item \emph{Cross-file consistency}: alternative encodings of the same data are isomorphic copies of each other.
\item \emph{Join validity}: every join by identifier between files agrees with the join by content (here, names).
\item \emph{Feature provenance}: no attribute available to a method is derived from held-out interactions.
\item \emph{Protocol sanity}: results exceed a random ranking, and the selected configurations converge within the budget.
\item \emph{Task shortcuts}: every derived task is compared with a baseline that reads the released text.
\end{enumerate}
Groups 1--3 run as one command over the \num{val.r.files} released files of the recommendation task, after checking them
against the pinned release revision: \num{val.r.checks} checks take \num{val.r.time}~s and \num{val.r.mem}~GB of memory in a
single process on a CPU. On the released files only the identity checks of the join fail, with \num{val.r.viol} violations in
the three misaligned KGs; on the corrected files all checks pass, and the correction leaves the KG triples, joined by entity
names, and the interaction files unchanged.
Figure~\ref{fig:pipeline} shows where in the benchmark's pipeline each issue arises, and Table~\ref{tab:issues}
summarizes how each is detected and its effect; this section describes the data issues, and Sections~\ref{sec:budget}
and~\ref{sec:prov} the protocol and task issues.

\begin{table*}[t]
  \caption{Issues found in the HuggingBench release, their detection, effect and handling}
  \label{tab:issues}
  \footnotesize
  \setlength{\tabcolsep}{4pt}
  \begin{tabular}{p{0.17\textwidth}p{0.2\textwidth}p{0.25\textwidth}p{0.3\textwidth}}
    \toprule
    Issue & Detection & Effect & Our handling \\
    \midrule
    Six encodings of the same interactions (two unnamed matrix sets, one file set per KG) & Join named file sets on names;
      compare degree profiles of unnamed files & Per-user results of models trained on different file sets are not
      comparable; item metadata cannot be attached & Name joins; permutations of the unnamed matrices by color refinement
      (Sec.~\ref{sec:numberings}); release the mappings \\
    Item--entity misalignment in \num{kg.misaligned.w} of \num{kg.released.w} external KGs & Join items and entities by
      name; assert item $i$ = entity $i$ & KG side information detached from its items; KG methods underestimated by up to
      \num{mis.maxpct.all} (Sec.~\ref{sec:fix}) & Corrected KG files; alignment validator \\
    100-epoch budget truncates training & Best epoch of the selected configuration at the end of the budget
      & \num{bud.cens.min}--\num{bud.cens.max} of \num{bud.cens.total} settings exhaust the budget, depending on the seed;
      \num{bud.trunc.impr} of \num{bud.trunc.pairs} truncated runs improve, the consistently truncated ones by up to
      \num{bud.maxpct} (Sec.~\ref{sec:budget})
      & Longer budget with unchanged early stopping; report the best epoch and flag it for a convergence check \\
    Crawl-time popularity counters may include held-out interactions & Compare counters with benchmark likes per split
      & Potential leakage; no gain as an additive prior for EASE (Sec.~\ref{sec:leak}) & Do not use; document \\
    Model cards and node attributes contain the targets of the provenance tasks & Search card text and node fields for the
      target & A card-reading baseline nearly matches the best reported tracing method; label-naming node attributes are a
      potential leakage channel for classification (Sec.~\ref{sec:prov}) & Report text baselines; mask identifiers \\
    \bottomrule
  \end{tabular}
\end{table*}

\subsection{Split integrity}
The three splits are disjoint (no user--item pair occurs in two splits, and no pair is repeated within a split), and every
user and item has at least \num{audit.minint} interactions, consistent with the 5-core filter. The per-user training share
lies between \num{audit.share.min} and \num{audit.share.max} (median \num{audit.share.med}), as expected from a per-user
60/20/20 split. \num{audit.coldtest.w} items have test likes but no training likes; they cannot be recommended by
collaborative methods, and their effect on the metrics is negligible.

\subsection{Six encodings of the same interactions}
\label{sec:numberings}
The release encodes the same interactions in \num{audit.num.sets.w} file sets: two sets of sparse matrices without user or
item names, used by general collaborative filtering (LightGCN and the contrastive models) and by social recommendation (DSL,
MHCN), and one set of text files per external KG, used by the KG-based models, each with lists of user and item names.
Every set has its own numbering: between any two KG file sets, at least \num{audit.num.items.diffmin} of the
\num{data.items} items have different identifiers, and no mapping is released. Joined on names, the
\num{audit.num.kgsets.w} KG file sets contain identical training, validation and test splits. For both unnamed matrix
encodings, we recovered the permutation to the 1-hop files by color refinement (the 1-dimensional Weisfeiler--Leman
algorithm~\cite{wlkernel,colorrefinement}) on the bipartite interaction graph with edges labeled by split. Under the
recovered mappings, all three splits are identical; only \num{perm.indist.users} users and \num{perm.indist.items} items are
structurally indistinguishable, so their assignment does not affect any metric. The mappings are needed to compare models
trained on different file sets user by user and to attach item metadata from HuggingKG to the unnamed matrices; they also
explain how the misalignment below could arise unnoticed, since every KG was exported with its own numbering.

\subsection{Item--entity misalignment in the external KGs}
\label{sec:misalign}
KG-based recommenders read the interaction file and the KG triples by identifier only, and assume that recommendation item
$i$ is KG entity $i$, i.e., that the first $n$ entity identifiers are reserved for the $n$ items, as in the KG datasets on
which these methods were developed~\cite{kgat}.
The released KG files include an item list (item name and identifier) and an entity list (entity name and identifier), so
the assumption can be checked by a join on names. It holds for the homogeneous graph, but not for the 1-hop, 2-hop and
publisher graphs: in the 1-hop and 2-hop graphs none of the \num{data.items} items has the same identifier as its entity,
and in the publisher graph only \num{kgalign.publish} do. Checks of single files or keys do not reveal this: every released
KG file set passes \num{val.checks} key and range checks (contiguous identifiers, unique names, disjoint splits, all
identifiers in the interaction and triple files within range, and every item identifier a legal entity identifier); only
the join on names fails.

Table~\ref{tab:kgaudit} measures what the recommenders actually see. We count the endpoints of triples that are
recommendation items according to the entity names (facts about items), how many of them reach the right item under the
identity assumption, and how many endpoints land on an item they do not describe. In the 1-hop and 2-hop graphs none of the
\num{audit.kg.1hop.ends} item endpoints reaches its item, and no item has a single correct fact; instead,
\num{audit.kg.1hop.wrong} (1-hop) and \num{audit.kg.2hop.wrong} (2-hop) endpoints attach facts to unrelated items. The
lost facts include all lineage relations between items (fine-tune, adapter, quantization and merge;
\num{audit.kg.1hop.lineage} item endpoints), which describe exactly the near-duplicate structure of the hub. For
example, the model SmolLM2-1.7B-Instruct is item \num{ex.smol.item} in the interaction files but entity \num{ex.smol.ent}
in the 1-hop graph, where it has \num{ex.smol.ntrip} triples, including a quantization edge to SmolVLM-Instruct (entity
\num{ex.vlm.ent}, also outside the item range). A KG recommender instead attaches to it entity \num{ex.smol.item}, the
Space \num{ex.smol.wrongname}, which has \num{ex.smol.nwrong.w} triple. In the publisher graph, \num{audit.kg.publish.ok} of
\num{audit.kg.publish.ends} item endpoints are correct. The KG-based results of the original study were therefore obtained
with side information that is, in effect, detached from the items.

We built corrected versions of the three graphs, in which every item keeps its interaction identifier as its entity
identifier and all other entities are renumbered after the items; all triples are kept. In the corrected files every item
endpoint reaches its item (Table~\ref{tab:kgaudit}). The same name join, run as an assertion before training, detects the
error in seconds.

\begin{table}[t]
  \caption{What KG recommenders see in the released and corrected KGs}
  \label{tab:kgaudit}
  \footnotesize
  \setlength{\tabcolsep}{3pt}
  \begin{tabular*}{\linewidth}{@{\extracolsep{\fill}}llrrrrr}
    \toprule
    KG & & Triples & Item ends & Correct & Wrong item & Items \\
    \midrule
1-hop & rel. & 1{,}180{,}911 & 767{,}927 & 0 & 117{,}260 & 0 \\
 & corr. & 1{,}180{,}911 & 767{,}927 & 767{,}927 & 0 & 25{,}080 \\
\midrule
2-hop & rel. & 3{,}063{,}081 & 767{,}927 & 0 & 99{,}312 & 0 \\
 & corr. & 3{,}063{,}081 & 767{,}927 & 767{,}927 & 0 & 25{,}080 \\
\midrule
Publisher & rel. & 25{,}078 & 25{,}078 & 4 & 39{,}535 & 4 \\
 & corr. & 25{,}078 & 25{,}078 & 25{,}078 & 0 & 25{,}078 \\
\bottomrule

  \end{tabular*}
  \tabnote{rel.: released; corr.: corrected; item $i$ = entity $i$ assumed. Item ends: triple endpoints that are recommendation items by entity name. Correct: item ends that reach their item. Wrong item: endpoints attached to an item they do not describe. Items: items with at least one correct fact.}
\end{table}

\subsection{Leakage hazards}
\label{sec:leak}
\paragraph*{Crawl-time counters} HuggingKG stores the number of likes and downloads of every model, dataset and Space at
crawl time. These counters are not recomputed from the training split and may contain the benchmark's held-out likes: for
\num{audit.cnt.ge} of the \num{data.items} items, the like counter is at least the item's number of likes in all three
splits, and it is more correlated with an item's test likes than the training likes are (Spearman \num{audit.sp.cnt} vs.\
\num{audit.sp.train}). As a popularity ranking it is not stronger than training popularity (Recall@10 \num{audit.pop.cnt}
vs.\ \num{audit.pop.train}), because it also counts likes from users outside the 5-core. Adding a validation-weighted
log-like-count prior to EASE did not improve Recall@10 (\num{leak.cnt.d} \num{leak.cnt.ci}), and subtracting the
benchmark's held-out likes from that counter had no measurable effect (cleaned minus raw: \num{leak.clean.d}
\num{leak.clean.ci}). This control is specific to the additive prior; we exclude crawl-time counters because their
provenance is not restricted to the training split, and recommend that releases either drop them or recompute them from
the training split.

\paragraph*{No interaction edges in the KGs} A KG that contains the user--item interactions themselves would leak test
likes into KG-based recommenders. The released external KGs contain \num{audit.rel.1hop} (1-hop and 2-hop),
\num{audit.rel.publish} (publisher) and \num{audit.rel.homo} (homogeneous) relation types, none of which is a like or another
user--item interaction of the benchmark, so this form of leakage does not occur.

\subsection{Identity across benchmark releases}
\label{sec:releases}
The misalignment is a failure that the KG format permits rather than an error of one release. The format used by HuggingBench
and SSLRec, inherited from KGAT~\cite{kgat}, stores item and entity names, but its loaders link items to entities only by the
convention that item $i$ is entity $i$, which holds by construction when items are numbered first and which none of the
loaders we examined (SSLRec and the original KGIN code) checks. We ran the same checks on other releases in this format
(Table~\ref{tab:releases}). Releases that number items first pass: the KGAT releases of Amazon-Book and
Yelp2018~\cite{kgat}, which KGIN distributes unchanged~\cite{kgin}, and their re-release with KGCL~\cite{kgcl}, which has the
same interactions and an exactly relabeled item--entity graph, have no identity violation; the only defects are formatting
ones, such as concatenated fields in one name list~\cite{kgatissue62}, and each check takes less than \num{ext.time.s}
seconds. Releases of one dataset need not share a numbering, however. KGPolicy~\cite{kgpolicy} re-releases Amazon-Book and
Yelp2018 with the same training interactions and KG facts that all occur in the KGAT release, but with different item
identifiers; MCCLK~\cite{mcclk} ships the MovieLens-1M files of MKR~\cite{mkr} (\num{xr.mcclk.ent} KG entities), while its
documentation reports the statistics of the RippleNet version~\cite{ripplenet} (\num{xr.mcclk.ent.doc} entities), whose item
identifiers differ. Each release is consistent on its own; the incompatibility arises when files of different releases are
combined. Loaders that infer the numbers of items and entities from the files, such as those of KGAT and KGIN, then raise no
error and detach the KG as in HuggingBench.

\begin{table}[t]
  \caption{Item identity within and across KG releases}
  \label{tab:releases}
  \footnotesize
  \setlength{\tabcolsep}{3pt}
  \begin{tabular*}{\linewidth}{@{\extracolsep{\fill}}llccc}
    \toprule
    Dataset & Releases & Within & Same ID & Kept \\
    \midrule
Amazon-Book & KGAT--KGCL & 0 / 0 & all 24{,}915 & 100\% \\
Amazon-Book & KGAT--KGPolicy & 0 / 0 & 32 of 24{,}915 & 0.7\% \\
Yelp2018 & KGAT--KGCL & 0 / 0 & all 45{,}538 & 100\% \\
Yelp2018 & KGAT--KGPolicy & 0 / 0 & 489 of 45{,}538 & 2.4\% \\
MovieLens-1M & MKR--RippleNet & 0 / 0 & 67 of 2{,}347 & -- \\
\bottomrule

  \end{tabular*}
  \tabnote{Within: violations of item $i$ = entity $i$ in each release (joined by name). Same ID: item identifiers that denote the same item in both releases. Kept: training pairs of the second release that keep their item when read with the item list of the first.}
\end{table}

\section{Replication of the Reported Results}
\label{sec:rep}
We reran all reported recommenders with the authors' configuration files and unmodified hyperparameter grids, fixed seeds
and early stopping. The learning rates listed in the grids are not applied by the framework's tuner, so all grid points
train with the default learning rate of each configuration file, and grid points that differ only in the learning rate
repeat the same run; we use these repetitions to measure run-to-run variation. Table~\ref{tab:rep} compares the results. LightGCN and DCCF match all \num{rep.lightgcn.nmatch.w} reported
Recall and NDCG values to four decimal places, and SimGCL, AutoCF, DSL, MHCN,
LightGCL, KGCL and KGRec replicate to within \num{rep.within.all} Recall@10 with every graph except KGCL with the 2-hop
graph (\num{rep.kgcl.2hop.d}). HCCF differs by \num{rep.hccf.absdelta.3}, within the spread of repeated runs of its selected
configuration (Recall@10 \num{rep.hccf.rep.lo}--\num{rep.hccf.rep.hi}), which nondeterministic GPU operations make diverge
after a few epochs; for LightGCN, SimGCL, AutoCF and LightGCL, the repetitions agree to within \num{rpt.lightgcl} in every
test metric. KGRec with three layers on the 2-hop graph---the configuration selected on validation data, which needs
a 32~GB GPU---reproduces the reported value (\num{rep.kgrec.2hop.r10} vs.\ \num{rep.kgrec.2hop.rep}). The SSLRec port of
KGIN replicates the reported near-zero scores (Recall@10 \num{kgin.rel.1hop.r10} with the 1-hop graph, the level of a random
ranking); we ran the port with this graph only and use the original implementation for KGIN elsewhere.
Table~\ref{tab:rep} follows the original selection, which re-evaluates the restored best state; our three-seed analysis
selects by the maximum of the validation curve, which for LightGCL with seed 42 gives Recall@10 \num{lgcl.s42.seeds} instead
of \num{lgcl.s42.rq1}. The reported
numbers are thus a faithful record of what the released code computes on the released data; the following sections examine
what they measure.

\begin{table}[t]
  \caption{Replication of the reported results}
  \label{tab:rep}
  \footnotesize
  \setlength{\tabcolsep}{3pt}
  \begin{tabular*}{\linewidth}{@{\extracolsep{\fill}}lccccc}
    \toprule
    & \multicolumn{2}{c}{Reported} & \multicolumn{2}{c}{Reproduced} & \\
    \cmidrule(lr){2-3}\cmidrule(lr){4-5}
    Method & R@10 & N@10 & R@10 & N@10 & $\Delta$R@10 \\
    \midrule
LightGCN & 0.1301 & 0.1003 & 0.1301 & 0.1003 & $+0.0000$ \\
HCCF & 0.1254 & 0.0975 & 0.1275 & 0.0988 & $+0.0021$ \\
SimGCL & 0.1515 & 0.1158 & 0.1515 & 0.1158 & $+0.0000$ \\
LightGCL & 0.1558 & 0.1198 & 0.1561 & 0.1198 & $+0.0003$ \\
AutoCF & 0.1530 & 0.1174 & 0.1529 & 0.1174 & $-0.0001$ \\
DCCF & 0.1493 & 0.1142 & 0.1493 & 0.1142 & $+0.0000$ \\
DSL & 0.1425 & 0.1099 & 0.1433 & 0.1099 & $+0.0008$ \\
MHCN & 0.1490 & 0.1154 & 0.1489 & 0.1158 & $-0.0001$ \\
\midrule
KGCL (homog.) & 0.1578 & 0.1220 & 0.1575 & 0.1222 & $-0.0003$ \\
KGCL (1-hop) & 0.1490 & 0.1160 & 0.1496 & 0.1165 & $+0.0006$ \\
KGCL (2-hop) & 0.1510 & 0.1170 & 0.1495 & 0.1159 & $-0.0015$ \\
KGCL (publisher) & 0.1543 & 0.1195 & 0.1537 & 0.1194 & $-0.0006$ \\
KGRec (homog.) & 0.0985 & 0.0751 & 0.0995 & 0.0755 & $+0.0010$ \\
KGRec (1-hop) & 0.0897 & 0.0681 & 0.0887 & 0.0678 & $-0.0010$ \\
KGRec (2-hop) & 0.0941 & 0.0729 & 0.0942 & 0.0726 & $+0.0001$ \\
KGRec (publisher) & 0.0941 & 0.0734 & 0.0946 & 0.0738 & $+0.0005$ \\
KGIN (1-hop) & 0.0004 & 0.0003 & 0.0004 & 0.0003 & $+0.0000$ \\
\bottomrule

  \end{tabular*}
  \tabnote{Test set. Our runs use seed 42, the seed of the released configuration files, and select the configuration on validation data with the original early-stopping criterion. KG methods: external graph in parentheses. $\Delta$: reproduced minus reported Recall@10.}
\end{table}

\section{Effect of Correcting the Knowledge Graphs}
\label{sec:fix}
Table~\ref{tab:mis} reruns every KG-based method with the corrected graphs, with the same grids and selection. With the
reported 100-epoch budget, the correction improves every method on every graph in every seed, by \num{mis.minpct.all} to
\num{mis.maxpct.all} in mean Recall@10; the gains are smallest for KGCL and largest for KGRec. For KGRec and the original
KGIN, every corrected graph also exceeds the homogeneous graph in every seed; for KGCL, the corrected graphs are within
\num{kgcl.homo.gap} of it on average, and the corrected publisher graph exceeds it in every seed. This changes the observed
ranking of graph variants: KGCL with the corrected publisher graph (Recall@10 \num{kgcl.publishfix.r10}) is the best KG-based
result overall. The corrections also change conclusions about individual methods: KGIN, near zero in the
original study, reaches \num{kgin.o3.1hopfix.4} with its original implementation and the corrected 1-hop graph, close to the
graph-based methods.

\paragraph*{Identity or kind of side information?} The correction changes two things for an item: it receives facts about
itself, and side information typical of items of its type and KG degree. We compare correct alignment with three random
assignments of the item entities of the corrected 1-hop graph, stratified by item type and coarse degree bins (0, 1, 2--4,
5--9, 10--19, 20--49 and at least 50 facts), training KGRec with two layers in each seed. The correctly aligned graph
outperforms every such assignment, by \num{perm.gap.100} Recall@10 after 100 epochs and \num{perm.gap.500} after 500 (means
over seeds), and every assignment outperforms the released graph, in all \num{perm.runs.w} runs. This control preserves the
graph-wide degree distribution but not each item's exact degree. A stricter assignment within item type and exact degree,
which moves \num{pex.items.pct}\% of the items but only \num{pex.ends.pct}\% of their KG endpoints, also stays below the
corrected graph in every seed, by \num{pex.gap.100} and \num{pex.gap.500} Recall@10, and is no better than the coarse
assignments. Neither control isolates semantic identity from all structural effects.

\paragraph*{Where the correction helps} Relative to the item's popularity, the correction helps most in the long tail.
With the 500-epoch budget and the 1-hop graph, KGRec (seed 42) retrieves in its top 10 only \num{kgbd.p1.rel} of the test likes of
items with 1--4 training likes when the graph is misaligned, and \num{kgbd.p1.fix} when it is corrected; for items with
20--99 training likes the share rises from \num{kgbd.p20.rel} to \num{kgbd.p20.fix}, whereas for the most popular items
(at least 100 training likes) it changes only from \num{kgbd.p100.rel} to \num{kgbd.p100.fix}. With the 2-hop graph,
popular items gain more in absolute terms (\num{kgbd2.p100.rel} to \num{kgbd2.p100.fix}), but the relative gain is again
far larger for the rarest items (\num{kgbd2.p1.rel} to \num{kgbd2.p1.fix}); all paired intervals exclude zero, and with the
1-hop graph the same pattern holds in seeds 43 and 44. Items with
fewer than 100 training likes receive \num{kgbd.longtail.share} of the test likes. With misaligned side information, a KG
recommender concentrates its hits on popular items, and the error hides in an aggregate metric that these items
dominate.

\paragraph*{Interaction with the budget} The measured correction gain also depends on the training budget: with 500 epochs
the correction still improves KGRec on every graph in every seed (Table~\ref{tab:mis}), and Section~\ref{sec:budget}
examines this interaction.

\paragraph*{How often the conclusions change} We count the pairs of settings compared in our tables whose order reverses
consistently, i.e., in all three seeds, and separate three causes. Correcting the KGs reverses \num{cc.a.flip.w} of the
\num{cc.a.pairs} orderings among KG configurations (variants of the graph for one method, and methods for one graph); for
example, the original KGIN leads KGRec on the released 2-hop graph and trails it on the corrected one. Replacing the
framework's port of KGIN with the original implementation reverses \num{cc.b.flip.w} of the \num{cc.b.pairs.w} orderings of
KGIN against the other reported methods, and extending the budget to 500 epochs (Section~\ref{sec:budget}) reverses
\num{cc.c.flip.w} of \num{cc.c.pairs} orderings. Of the remaining pairs, \num{cc.a.stable.w}, \num{cc.b.stable.w} and
\num{cc.c.stable} keep their order in every seed, and the others are not consistent across seeds.

\begin{table*}[t]
  \caption{Effect of correcting the external KGs (test Recall@10)}
  \label{tab:mis}
  \small
  \begin{tabular*}{\linewidth}{@{\extracolsep{\fill}}lccccccc}
    \toprule
    & Homogeneous & \multicolumn{2}{c}{1-hop} & \multicolumn{2}{c}{2-hop} & \multicolumn{2}{c}{Publisher} \\
    \cmidrule(lr){3-4}\cmidrule(lr){5-6}\cmidrule(lr){7-8}
    Method & & rel. & corr. & rel. & corr. & rel. & corr. \\
    \midrule
KGCL & 0.1585 & 0.1496 & 0.1591 & 0.1503 & 0.1581 & 0.1540 & 0.1604 \\
KGRec & 0.0987 & 0.0845 & 0.1457 & 0.0937 & 0.1479 & 0.0943 & 0.1204 \\
KGRec, 500 epochs & 0.0987 & 0.1107 & 0.1481 & 0.0969 & 0.1493 & 0.1054 & 0.1379 \\
KGIN (original code) & 0.1351 & 0.1169 & 0.1445 & 0.1198 & 0.1446 & 0.1152 & 0.1475 \\
\bottomrule

  \end{tabular*}
  \tabnote{Mean over three seeds; configuration selected on validation data in each seed. Every corrected value exceeds the released one in every seed. rel.: released graph; corr.: corrected graph. The homogeneous graph is aligned in the release and is shown for reference. KGIN: the authors' original implementation. KGRec, 500 epochs: same grid with a 500-epoch budget and unchanged early stopping (Section~\ref{sec:budget}); its homogeneous runs stop early under both budgets.}
\end{table*}

\section{The Training Budget}
\label{sec:budget}
\paragraph*{Which runs are truncated?} Early stopping is meant to make the number of epochs irrelevant, but only if the
budget is large enough for it to trigger. We recorded, for every setting we ran with the 100-epoch budget
(\num{bud.cens.total} method--graph combinations, released and corrected graphs) and every seed, the epoch at which the
selected configuration reached its best validation value and why it stopped. In \num{bud.cens.s42.w}, \num{bud.cens.s43.w}
and \num{bud.cens.s44.w} settings of seeds 42, 43 and 44, the run reached the 100-epoch limit before early stopping.
\num{bud.cens.all.n.w.cap} settings reach the limit in every seed: LightGCN, DSL, and KGRec with the released 1-hop, released
publisher and corrected publisher graphs; LightGCL, the original KGIN with the released 1-hop graph and KGRec with the
corrected 1-hop and both 2-hop graphs reach it only in some seeds (Table~\ref{tab:budget}).

\begin{figure}[t]
  \centering
  \includegraphics[width=\columnwidth]{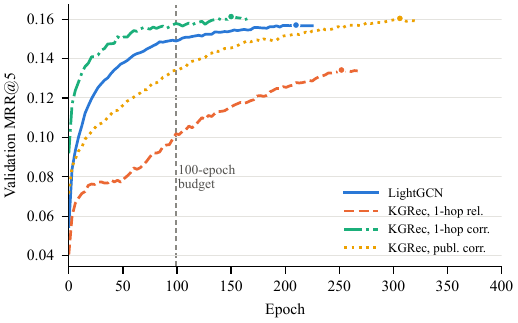}
  \caption{Validation MRR@5 (the early-stopping criterion) during training with a 500-epoch budget (seed 42); the dashed
  line marks the reported 100-epoch budget, dots the best epoch.}
  \label{fig:curves}
\end{figure}

\paragraph*{Extending the budget} In seeds 43 and 44 every run used a budget of 500 epochs; in seed 42 we reran every
setting whose selected configuration reached the 100-epoch limit with 500 epochs and otherwise identical configurations and
early stopping, and in all other settings of seed 42 every grid configuration stopped early, so their result is the same
under both budgets. A hook in the training loop records, within the same run, the test
result of the best state after 100 epochs, so both budgets are evaluated on the same trajectory; short runs with and without
the hook confirmed that it does not change training beyond the GPU nondeterminism already present before it fires.
Of the \num{bud.trunc.pairs} setting--seed pairs stopped by the limit, \num{bud.trunc.impr} improve under the longer budget,
and not by a few extra epochs: their best epochs lie between \num{bud.best.min} and \num{bud.best.max}. Reaching the limit
does not guarantee a better test result: DSL in seed 44 keeps the same selected state and score (it stops early at epoch
\num{bud.dsl44.stop}, best epoch \num{bud.dsl44.best}), and LightGCL in seed 43 improves on validation but slightly decreases on
test (\num{bud.lgcl43.d}). \num{bud.cens.allimpr.n.w.cap} of the settings that reach the limit in every seed improve in every
seed: LightGCN gains \num{bud.lgcn.d} Recall@10 (\num{bud.lgcn.rng} across seeds), and KGRec with the released 1-hop graph
\num{bud.kgrec1hop.d} (+\num{bud.kgrec1hop.pct}); settings that reach it only in some seeds gain little, and nothing in some
seeds.
Figure~\ref{fig:curves} shows typical validation curves: at epoch 100 they are still rising.

\begin{table}[t]
  \caption{Extending the 100-epoch budget to 500 epochs}
  \label{tab:budget}
  \scriptsize
  \setlength{\tabcolsep}{2.5pt}
  \begin{tabular*}{\linewidth}{@{\extracolsep{\fill}}lcccl}
    \toprule
    Setting & Trunc. & 100 ep. & 500 ep. & $\Delta$R@10 [min, max] \\
    \midrule
LightGCN & 3 & 0.1297 & 0.1362 & $+0.0065$ [$+0.0048$, $+0.0076$] \\
DSL & 3 & 0.1433 & 0.1456 & $+0.0023$ [$0.0000$, $+0.0041$] \\
LightGCL & 1 & 0.1401 & 0.1400 & $0.0000$ [$-0.0001$, $0.0000$] \\
KGIN (orig.), 1-hop rel. & 1 & 0.1169 & 0.1171 & $+0.0002$ [$0.0000$, $+0.0005$] \\
KGRec, 1-hop rel. & 3 & 0.0845 & 0.1107 & $+0.0262$ [$+0.0215$, $+0.0291$] \\
KGRec, 1-hop corr. & 2 & 0.1457 & 0.1481 & $+0.0024$ [$0.0000$, $+0.0044$] \\
KGRec, 2-hop rel. & 2 & 0.0937 & 0.0969 & $+0.0032$ [$0.0000$, $+0.0090$] \\
KGRec, 2-hop corr. & 2 & 0.1479 & 0.1493 & $+0.0014$ [$0.0000$, $+0.0021$] \\
KGRec, publ.\ rel. & 3 & 0.0943 & 0.1054 & $+0.0111$ [$+0.0100$, $+0.0126$] \\
KGRec, publ.\ corr. & 3 & 0.1204 & 0.1379 & $+0.0175$ [$+0.0118$, $+0.0216$] \\
\bottomrule

  \end{tabular*}
  \tabnote{Settings whose selected configuration reaches the 100-epoch limit in at least one seed; unchanged early stopping; test Recall@10, mean over three seeds; configuration selected on validation data under each budget. Trunc.: number of seeds in which the setting reaches the limit. $\Delta$: mean difference, computed before rounding, with its range over seeds; a zero marks a seed in which the longer budget leaves the result unchanged.}
\end{table}

\paragraph*{What changes} The budget affects two conclusions. First, it inflates the measured cost of the misalignment: for
KGRec with the 1-hop graph, the gap between the corrected and the released graph shrinks in every seed, from
\num{bud.gap1hop.100} to \num{bud.gap1hop.500} Recall@10 on average, because the model trained on the misaligned graph
converges more slowly. With the model depth fixed, the same holds for three layers: extending the budget reduces the
correction gain in every seed, by \num{fd.L3.red.min} to \num{fd.L3.red.max} Recall@10 (paired over users, all intervals
excluding zero). With two layers, the run on the released graph stops early in seeds 43 and 44 (best epochs
\num{fd.L2.rel.ep.hi} and \num{fd.L2.rel.ep.lo}), so only the corrected graph gains from the longer budget and the change in
the correction gain is positive (\num{fd.L2.inter.o.lo} and \num{fd.L2.inter.o.hi}, against \num{fd.L2.inter.s42} in seed
42). In all six cases the corrected graph remains better under both budgets, and the correction remains large and consistent
(Section~\ref{sec:fix}). Second, the budget
changes the differences between KG-based methods: with 500 epochs, KGRec with the corrected publisher graph improves from
\num{bud.kgrecpubfix.100} to \num{bud.kgrecpubfix.500}, which closes most of its gap to the original KGIN with the same
graph (\num{kgin.o3.publishfix}). The ranking of the strongest methods does not change: the best
graph-based and KG-based results stay below the simple baselines of Section~\ref{sec:side}; LightGCN, for example, moves
from \num{bud.lgcn.100} to \num{bud.lgcn.500}.

\paragraph*{Learning rates} The budget also interacts with the learning rate: smaller learning rates need more epochs, so a
100-epoch budget can make them look worse than they are. We therefore trained LightGCL, the strongest reported general
recommender, with learning rates $10^{-4}$ and $10^{-5}$ set directly in the optimizer configuration and the 500-epoch
budget. In every seed its best validation value remains below that of the default ($10^{-3}$), by at least
\num{lr.lgcl.margin.min}, so validation would still select the default, and with $10^{-5}$ even 500 epochs are not enough to
converge (best epoch \num{lr.lgcl.5.ep} or later). For LightGCL, the learning-rate
controls thus do not improve the validation-selected result. For KGCL with the corrected publisher graph, the best KG-based
configuration, we added learning rates $10^{-2}$ and $10^{-4}$ for both depths: with 100 epochs validation still selects the
default in every seed, and with 500 epochs it selects $10^{-4}$ with three layers, which raises Recall@10 from
\num{r7.def.500} to \num{r7.r10.500}, below EASE (\num{ease.r10}). The other comparisons retain the released effective
learning rates and selection procedures.

\section{Simple Baselines and Unused Side Information}
\label{sec:side}
\paragraph*{Linear models} Table~\ref{tab:main} compares the reported, reproduced and corrected results with simple
baselines, tuned on a fixed quarter of the validation users: ItemKNN with cosine similarity and shrinkage, RP3beta~\cite{rp3beta},
EASE~\cite{ease} ($\lambda=\num{ease.lambda}$) and EASE with data-adaptive normalization (DAN~\cite{dan}). EASE, a closed-form
item--item model with a single L2 penalty, reaches Recall@10 \num{ease.r10}, above every reported, reproduced and corrected
result; paired with the reproduced LightGCN (seed 42) on the same test users, it gains \num{ease.vs.lgcn} \num{ease.vs.lgcn.ci}.
EASE-DAN is the strongest purely collaborative model (Recall@10 \num{dan.r10}, \num{dan.vs.ease} \num{dan.vs.ease.ci} over
EASE), \num{dan.vs.bestrep.pct} above the best reported result. HuggingBench's data are sparse and long-tailed, with many
near-duplicate items (quantizations and adapters of the same base model); on such data item--item models are hard to beat.
Within the configurations, selection criteria and training budgets evaluated here, the linear models achieve the highest
observed scores.

\begin{table}[t]
  \caption{Resource recommendation (test, full ranking)}
  \label{tab:main}
  \footnotesize
  \setlength{\tabcolsep}{2.5pt}
  \begin{tabular*}{\linewidth}{@{\extracolsep{\fill}}llcccc}
    \toprule
    & Method & R@10 & R@20 & R@40 & N@10 \\
    \midrule
\multirow{4}{*}{Reported}  & LightGCN & 0.1301 & 0.1932 & 0.2759 & 0.1003 \\
 & LightGCL & 0.1558 & 0.2228 & 0.3017 & 0.1198 \\
 & KGCL (homo) & 0.1578 & 0.2237 & 0.3059 & 0.1220 \\
 & KGIN (homo) & 0.0096 & 0.0146 & 0.0219 & 0.0076 \\
\midrule
\multirow{4}{*}{Reproduced}  & LightGCN & 0.1301 & 0.1932 & 0.2759 & 0.1003 \\
 & LightGCL & 0.1561 & 0.2229 & 0.3018 & 0.1198 \\
 & KGCL (homo) & 0.1575 & 0.2254 & 0.3066 & 0.1222 \\
 & KGIN (1-hop) & 0.0004 & 0.0009 & 0.0017 & 0.0003 \\
\midrule
\multirow{2}{*}{Corrected}  & KGIN (orig., 1-hop corr.) & 0.1445 & 0.2123 & 0.2984 & 0.1090 \\
 & KGCL (publisher corr.) & 0.1604 & 0.2291 & 0.3113 & 0.1242 \\
\midrule
\multirow{2}{*}{500 ep.}  & LightGCN & 0.1362 & 0.2018 & 0.2857 & 0.1048 \\
 & KGRec (publisher, corr.) & 0.1379 & 0.2018 & 0.2847 & 0.1058 \\
\midrule
\multirow{5}{*}{Ours: CF}  & Popularity & 0.0358 & 0.0653 & 0.1049 & 0.0265 \\
 & ItemKNN & 0.1406 & 0.1962 & 0.2660 & 0.1116 \\
 & RP3beta & 0.1699 & 0.2364 & 0.3172 & 0.1319 \\
 & EASE & 0.1684 & 0.2328 & 0.3076 & 0.1332 \\
 & EASE-DAN & 0.1746 & 0.2400 & 0.3149 & 0.1376 \\
\midrule
\multirow{3}{*}{\shortstack[l]{Ours: CF +\\side info}}  & EASE + item graph & 0.1776 & 0.2441 & 0.3221 & 0.1395 \\
 & EASE + TF-IDF text & 0.1846 & 0.2540 & \textbf{0.3335} & 0.1453 \\
 & EASE + TF-IDF + graph & \textbf{0.1860} & \textbf{0.2551} & 0.3332 & \textbf{0.1470} \\
\bottomrule

  \end{tabular*}
  \tabnote{Reported: original paper. Reproduced: our runs of the authors' code, seed 42 as in Table~\ref{tab:rep}. Corrected: corrected KG (KGIN: original implementation). 500 ep.: 500-epoch budget. Corrected and 500 ep.: mean over three seeds. Ours: methods added in this study. Best value in bold.}
\end{table}

\paragraph*{Item text} For every item we extract from HuggingKG its type, identifier (organization/name), pipeline tag,
tags, libraries, languages and description (model or dataset card, cut to 3{,}000 characters; Spaces have none); the YAML
front matter of the cards is already stripped in the release. We deliberately do not use the crawl-time popularity
counters (Section~\ref{sec:leak}). We add this text to EASE by adapting the
closed-form side-information model of Jeunen et al.~\cite{jeunen2020}: with L2-normalized TF-IDF vectors $f_i$, we add a
weighted item--item similarity matrix with a zero diagonal, $S_{ij}=\cos(f_i,f_j)$ for $i\neq j$ and $S_{ii}=0$, to the Gram
matrix: $P=(X^\top X + wS + \lambda I)^{-1}$, $B = I - P\,\mathrm{diag}(P)^{-1}$, $\mathrm{diag}(B)=0$. With $w$ selected on
validation data from $\num{grid.text}$, Recall@10 rises to \num{tf.r10} (\num{tf.vs.ease} \num{tf.vs.ease.ci} over EASE);
additive fusions of EASE with text scores gain only \num{add.gain.profile} and \num{add.gain.knn}. Adding the item--item
graph of HuggingKG (the seven relations of the homogeneous graph plus shared publisher) in the same closed form gives
Recall@10 \num{tfg.r10}, a small but significant further gain (\num{tfg.vs.tf} \num{tfg.vs.tf.ci}); text alone exceeds the
best reported result by \num{tf.vs.bestrep.pct}, and text with the graph by \num{tfg.vs.bestrep.pct}. The released KG, once correctly joined, thus helps; the item
text helps much more.

\paragraph*{Representation} Table~\ref{tab:text} compares TF-IDF with dense embeddings from bge-m3~\cite{bgem3} and
Qwen3-Embedding-0.6B~\cite{qwen3emb} under matched conditions: the same input text (descriptions cut to 6{,}000 characters,
then truncated by each encoder's tokenizer to 1{,}024 tokens, with TF-IDF computed on exactly the text the encoder sees),
the same closed form, both as a full similarity matrix and as a symmetrized 100-nearest-neighbor graph, and weights selected
on validation data from grids whose optimum lies inside. TF-IDF wins all four comparisons by \num{mt.diff.min} to
\num{mt.diff.max} Recall@10, with all intervals excluding zero. Hub text is highly templated: the median TF-IDF cosine of two
models by the same publisher is \num{pub.cos} (\num{pub.dup} near-duplicates), compared with \num{rnd.cos} for random pairs.
Which fields carry the signal? With the same closed form and one weight grid for all views, the item type and identifier
alone (organization and name) score close to the full text (Recall@10 \num{fld.id.r10}; \num{fld.id.vs.full} \num{fld.id.vs.full.ci}),
and remain close with the graph added as in the best model (\num{fldg.id.r10} against \num{tfg.r10}; \num{fldg.id.vs.full}
\num{fldg.id.vs.full.ci}),
the description alone gains \num{fld.desc.vs.ease} over EASE, and still \num{fld.mask.vs.ease} after URLs, repository
identifiers and the item's own organization and name are masked (\num{fld.mask.n} of \num{fld.desc.n} descriptions change);
the type and remaining metadata (pipeline tag, tags, libraries, languages and further tags) gain \num{fld.meta.vs.ease}. All
five views improve over EASE, with paired intervals excluding zero. The signal is thus redundant across fields: identifiers,
which encode the publisher and the model family, carry it in compact form, and descriptions remain useful after the mask,
which does not remove all model-family references or indirect identity cues.

\begin{table}[t]
  \caption{Text representations under matched conditions (test Recall@10)}
  \label{tab:text}
  \footnotesize
  \setlength{\tabcolsep}{2.6pt}
  \begin{tabular*}{\linewidth}{@{\extracolsep{\fill}}lllcc}
    \toprule
    View & Similarity & Repr. & R@10 & $\Delta$R@10 [95\% CI] \\
    \midrule
\multirow{4}{*}{bge-m3} & \multirow{2}{*}{full} & TF-IDF & 0.1846 & -- \\
 & & dense & 0.1799 & $-0.0047$ [$-0.0059$, $-0.0034$] \\
 & \multirow{2}{*}{kNN-100} & TF-IDF & 0.1844 & -- \\
 & & dense & 0.1780 & $-0.0064$ [$-0.0077$, $-0.0049$] \\
\midrule
\multirow{4}{*}{Qwen3-Emb.} & \multirow{2}{*}{full} & TF-IDF & 0.1847 & -- \\
 & & dense & 0.1801 & $-0.0046$ [$-0.0059$, $-0.0033$] \\
 & \multirow{2}{*}{kNN-100} & TF-IDF & 0.1844 & -- \\
 & & dense & 0.1767 & $-0.0077$ [$-0.0090$, $-0.0062$] \\
\bottomrule

  \end{tabular*}
  \tabnote{View: the encoder whose tokenizer truncates the text; TF-IDF uses the same text. Full: full cosine-similarity matrix; $k$NN-100: symmetrized 100-nearest-neighbor graph. $\Delta$: dense minus TF-IDF with 95\% CI.}
\end{table}

\paragraph*{Where text helps} Table~\ref{tab:breakdown} breaks the results down by item popularity, item type and user
activity. Text and graph mainly help rarely liked items: for items with 1--4 training likes the share of test likes retrieved
in the top 10 rises from \num{bd.p1.ease} (EASE) to \num{bd.p1.tfg}, whereas for the \num{bd.p100.items} most popular items it
changes little (\num{bd.p100.d} \num{bd.p100.ci}). Models gain most (\num{bd.model.d}) and Spaces, which have no card, least
(\num{bd.space.d}); the gain decreases with user activity.

\begin{table}[t]
  \caption{Results by item and user group}
  \label{tab:breakdown}
  \footnotesize
  \setlength{\tabcolsep}{2.5pt}
  \begin{tabular*}{\linewidth}{@{\extracolsep{\fill}}lccccc}
    \toprule
    Group & Share & LGCN & EASE & +T+G & $\Delta$ \\
    \midrule
    \multicolumn{6}{l}{\emph{Items by training likes}} \\
1--4 & 0.048 & 0.0059 & 0.0057 & 0.0339 & $+$0.0282 \\
5--19 & 0.186 & 0.0122 & 0.0156 & 0.0407 & $+$0.0250 \\
20--99 & 0.305 & 0.0128 & 0.0249 & 0.0405 & $+$0.0156 \\
$\ge$100 & 0.461 & 0.1579 & 0.1976 & 0.1949 & $-$0.0027 \\
\midrule
\multicolumn{6}{l}{\emph{Items by type}} \\
Model & 0.627 & 0.0808 & 0.1049 & 0.1168 & $+$0.0119 \\
Dataset & 0.094 & 0.0368 & 0.0548 & 0.0624 & $+$0.0076 \\
Space & 0.279 & 0.0903 & 0.1112 & 0.1161 & $+$0.0049 \\
\midrule
\multicolumn{6}{l}{\emph{Users by training likes}} \\
1--4 & 0.269 & 0.1570 & 0.2059 & 0.2287 & $+$0.0228 \\
5--9 & 0.372 & 0.1487 & 0.1904 & 0.2092 & $+$0.0187 \\
10--19 & 0.188 & 0.1108 & 0.1438 & 0.1600 & $+$0.0162 \\
20--49 & 0.116 & 0.0797 & 0.1032 & 0.1140 & $+$0.0108 \\
$\ge$50 & 0.055 & 0.0455 & 0.0589 & 0.0618 & $+$0.0029 \\
\bottomrule

  \end{tabular*}
  \tabnote{Share of test likes retrieved in the top 10 by item group, and mean Recall@10 by user group (groups by number of training likes). Share: fraction of test likes (items) or test users. LGCN: LightGCN (seed 42); +T+G: EASE + TF-IDF + graph; $\Delta$: +T+G minus EASE (all 95\% intervals exclude zero).}
\end{table}

\paragraph*{Cost} Fitting EASE takes \num{time.ease.s}~s on an 8-core CPU, dominated by one inversion of a
$\num{data.items}\times\num{data.items}$ matrix, and the best model with text and graph \num{time.best.s}~s including the
construction of both similarity matrices. A single configuration of LightGCN trains for about \num{time.lightgcn.min} minutes
on a GPU with the 100-epoch budget, and one of the original KGIN for up to \num{time.kgin.onehop.h} hours with the 1-hop
graph; a full grid of KGRec on a 2-hop graph with 500 epochs takes more than a GPU-day.

\section{Provenance: New Derivatives, Lineage and Tasks}
\label{sec:prov}
\subsection{Inheritance cold start}
New models on a hub typically appear as derivatives of existing ones and receive no interactions for a while. Lineage has
been used to predict the benchmark performance of such models~\cite{tamura2025crow}; here it serves to predict user interest.
We simulate
this with an \emph{inheritance cold-start} protocol on the released split. Among the \num{cold.withparent} recommendation
items with a parent item via a lineage relation (fine-tune, quantization, adapter, merge), we sample \num{cold.share.pct}
(\num{cold.new}) with a fixed seed as new models and remove all their training interactions (\num{cold.removed} likes). Half
of them select fusion weights, the other half (\num{cold.test} models) are used for testing: the held-out positives of a user
are all of the user's interactions with these models in the original splits, and every user with at least one positive
(\num{cold.users}) ranks the test models only. Text profiles score a new model by the cosine between the user's profile
(the normalized sum of the vectors of the user's training items) and the model's text vector; lineage scores combine the
user's interactions with, and EASE scores of, its parent models, and a typed variant also weights relation types and
aggregates the warm siblings of the new model, with weights selected on the validation half; combinations sum z-normalized
scores.

Table~\ref{tab:cold} shows that text is the dominant signal (TF-IDF profile: Recall@10 \num{cold.tfidf.r10}), lineage alone
reaches \num{cold.parents.r10}, and combining TF-IDF and lineage improves over either alone (\num{cold.tflin.r10};
\num{cold.tflin.vs.tf} \num{cold.tflin.vs.tf.ci}), so the two signals are complementary for this task; the texts of new
models are also not copies of their parents' texts (median parent--child TF-IDF cosine between \num{pc.cos.lo} and
\num{pc.cos.hi} across relation types). When combined with TF-IDF, the more elaborate typed variant brings no measurable
gain (\num{cold.typed.vs.untyped} \num{cold.typed.vs.untyped.ci}).

\begin{table}[t]
  \caption{Simulated inheritance cold start}
  \label{tab:cold}
  \footnotesize
  \setlength{\tabcolsep}{4pt}
  \begin{tabular*}{\linewidth}{@{\extracolsep{\fill}}lccc}
    \toprule
    Method & R@10 & N@10 & $\Delta$R@10 \\
    \midrule
Random & 0.0287 & 0.0152 &  \\
Lineage (parents) & 0.1863 & 0.1131 & $-0.1366$ \\
Text profile: Qwen3-Embedding-0.6B & 0.1343 & 0.0854 & $-0.1887$ \\
Text profile: bge-m3 & 0.2427 & 0.1531 & $-0.0803$ \\
Text profile: TF-IDF & 0.3229 & 0.2371 & -- \\
EASE + TF-IDF (closed form) & 0.3075 & 0.2261 & $-0.0155$ \\
TF-IDF + lineage & 0.3541 & \textbf{0.2546} & $+0.0311$ \\
TF-IDF + bge-m3 + lineage & \textbf{0.3602} & 0.2543 & $+0.0372$ \\
\bottomrule

  \end{tabular*}
  \tabnote{\num{cold.test} test models; each user ranks only these models. $\Delta$: Recall@10 difference to the TF-IDF profile.}
\end{table}

\subsection{Lineage tracing}
The tracing task asks for the base model of a derived model (\num{trace.n} test triples, \num{trace.finetune.pct}
fine-tunes); the reported methods are structure-only KG embeddings, with TransE~\cite{transe} best at MRR \num{tr.transe.rep}.
We replicated TransE with the authors' LibKGE~\cite{libkge} configuration and learning-rate grid; validation selects the same
learning rate in all three seeds, and the selected models reach a filtered MRR of \num{tr.transe.mrr} on average
(\num{tr.transe.mrr.lo}--\num{tr.transe.mrr.hi}; Hits@1 \num{tr.transe.h1}). Although the YAML front matter of the cards (which
may contain a \texttt{base\_model} field) is stripped in the release, the card bodies still name their base models. Such
references have been mined from cards and metadata to reconstruct the supply chain of the hub~\cite{hugginggraph}; here we
measure how much of the benchmark task they solve. A
training-free baseline that extracts the model identifiers mentioned in the derived model's card and ranks them by order of
appearance finds the correct base model among its candidates for \num{tr.mention.pct} of the test triples; assigning zero
reciprocal rank when the target is not extracted, it reaches MRR \num{tr.card.mrr} (Hits@1 \num{tr.card.h1}), close to
TransE without using the graph. Ranking the extracted models first and all other entities by TransE yields MRR
\num{tr.hyb.mrr} on average, \num{tr.hyb.vs.transe.lo} to \num{tr.hyb.vs.transe.hi} over TransE, with paired intervals over
test triples that exclude zero in every seed. The task thus partly measures how well a
method reads the card, a form of the metadata shortcuts that masking analyses also address in artifact-graph
prediction~\cite{artifactlinker}; future versions should report such text baselines or mask identifiers in the cards.

\subsection{Task classification}
The released classification test set contains \num{cls.test.n} items (the original paper reports \num{cls.test.reported}).
A literal matcher that predicts the task names occurring in an item's card reaches micro-F1 \num{cls.all.f1lit}, close to
predicting the most frequent task (\num{cls.all.f1maj}) and far below the best reported \num{cls.best.rep}; this weak matcher
does not rule out other textual shortcuts. A structured attribute of the item node---the \texttt{pipeline\_tag} of models
and the task fields of datasets---names one of the target labels for \num{cls.all.field} of the test items. This is a
potential leakage channel: the code that builds the released node features is not public, so we could not verify whether
the original classifiers use these attributes; a release should state it explicitly.

\section{Lessons for Benchmark and Model-Lake Builders}
\label{sec:lessons}
\paragraph*{Validate joins, not only files} Each released KG file set of HuggingBench passes key and range checks
(Section~\ref{sec:misalign}); the error lies in the join between files. Joins between interaction data and side information should be validated by content (here, entity
names) before any model is trained, and the validator should ship with the benchmark. The same holds for alternative file
sets of the same data, which should come with an explicit mapping.

\paragraph*{Prefer explicit links to implicit conventions} The item--entity convention fails silently both within a release
(Section~\ref{sec:misalign}) and across releases of one dataset (Section~\ref{sec:releases}). Formats whose loaders read an
explicit link avoid this failure: RecBole~\cite{recbole}, for example, requires a separate item--entity link file for
knowledge-aware models. Benchmarks and model lakes that join artifacts with side information should store such links
explicitly, or ship a validator that asserts the convention, and identify each release by a version and checksums rather than
by the name of the dataset~\cite{datasetversioning}.

\paragraph*{Release the provenance of every feature} The crawl-time counters and the label-naming node attributes are
correct data that are dangerous in the wrong place. A benchmark built from a live hub should record how and when every
attribute was obtained, and drop or recompute from the training split any attribute that may contain held-out
interactions.

\paragraph*{Make the protocol self-checking} A budget that binds the selected configuration is visible in the training
logs. Benchmarks should report the best epoch of every selected configuration and flag a best epoch at the budget for a
convergence check, and results at the level of a random ranking (Recall@10 \num{rnd.r10} here), such as the near-zero KGIN
scores of the original study, should be treated as errors rather than as findings.

\paragraph*{Include baselines that use what is released} Benchmarks that release side information should report at least
one simple model that uses it, and a tuned closed-form linear model (EASE or EASE-DAN) is a cheap, strong default for sparse
implicit feedback.

\paragraph*{Separate provenance from content} On a model hub, documentation, identifiers and lineage are intertwined:
cards name base models and tasks. Provenance tasks should mask the information they ask for, or report text baselines that
show how much of the task is solved by reading.

\paragraph*{Limitations} Likes are a weak form of feedback; the data are one snapshot with a random per-user split, and our
cold-start protocol simulates new items on this split rather than following models as they are published. Our seeds vary
the training of the models but not the split; we tuned learning rates explicitly only for LightGCL and KGCL and used
text encoders off the shelf.

\section{Conclusion}
\label{sec:artifacts}
HuggingBench's reported results are reproducible, but what they measure is shaped by properties of the release that no
result table shows: misaligned identifiers that detach the external KGs from their items, a training budget that ends
before several methods converge, and rich item text that no baseline uses. Correcting the KG joins consistently reverses
\num{cc.a.flip.w} of \num{cc.a.pairs} comparisons between KG configurations and changes the verdict on KGIN; the
implementation and the training budget affect further comparisons; and, within the configurations and selection procedures
evaluated here, closed-form models with item text achieve the highest observed scores. The
corrected KG files, the mappings between the file sets, the validators, the inheritance cold-start split, the patches,
scripts and logs of every run are available at \url{https://doi.org/10.5281/zenodo.23857277}.

\section*{Acknowledgment}
This manuscript acknowledges the use of Claude Opus 5.5~\cite{opus55}, developed by Anthropic, and of GPT-6.1
Sol~\cite{gpt61sol}, developed by OpenAI, to improve language clarity, refine sentence structure, and enhance overall writing
precision. The study design, experiments, analysis, results, and conclusions remain the sole responsibility of the authors.

\bibliographystyle{IEEEtran}
\bibliography{refs}

\end{document}